\documentclass[letterpaper]{article} 
\usepackage{aaai23}  
\usepackage{times}  
\usepackage{helvet}  
\usepackage{courier}  
\usepackage[hyphens]{url}  
\usepackage{graphicx} 
\usepackage{natbib}  
\usepackage{caption} 
\usepackage{algorithm}
\usepackage{algorithmic}

\usepackage{subfigure}
\usepackage{amsmath}
\usepackage{amssymb}
\usepackage{booktabs}
\usepackage{multirow}
\usepackage{dsfont}
\usepackage[colorlinks=true,
	linkcolor=red,
	filecolor=red,      
	urlcolor=magenta,
	citecolor=green,]{hyperref}

\usepackage{newfloat}
\usepackage{listings}
\DeclareCaptionStyle{ruled}{labelfont=normalfont,labelsep=colon,strut=off} 
\floatstyle{ruled}
\newfloat{listing}{tb}{lst}{}
\floatname{listing}{Listing}
\title{Superpoint Transformer for 3D Scene Instance Segmentation}
\author{
    Jiahao Sun\textsuperscript{\rm 1}, 
    Chunmei Qing\textsuperscript{\rm 1}\thanks{Corresponding author: Chunmei Qing}, 
    Junpeng Tan\textsuperscript{\rm 1}, 
    Xiangmin Xu\textsuperscript{\rm 2},
}
\affiliations{
    \textsuperscript{\rm 1} School of Electronic and Information Engineering, South China University of Technology, China\\
    \textsuperscript{\rm 2} School of Future Technology, South China University of Technology, China\\

    eesjh@mail.scut.edu.cn, qchm@scut.edu.cn, tjeepscut@gmail.com, xmxu@scut.edu.cn
}

\usepackage{bibentry}

\begin{document}

\maketitle

\begin{abstract}
Most existing methods realize 3D instance segmentation by extending those models used for 3D object detection or 3D semantic segmentation. However, these non-straightforward methods suffer from two drawbacks: 1) Imprecise bounding boxes or unsatisfactory semantic predictions limit the performance of the overall 3D instance segmentation framework. 2) Existing methods require a time-consuming intermediate step of aggregation. To address these issues, this paper proposes a novel end-to-end 3D instance segmentation method based on Superpoint Transformer, named as \textbf{SPFormer}. It groups potential features from point clouds into superpoints, and directly predicts instances through query vectors without relying on the results of object detection or semantic segmentation. The key step in this framework is a novel query decoder with transformers that can capture the instance information through the superpoint cross-attention mechanism and generate the superpoint masks of the instances. Through bipartite matching based on superpoint masks, SPFormer can implement the network training without the intermediate aggregation step, which accelerates the network. Extensive experiments on ScanNetv2 and S3DIS benchmarks verify that our method is concise yet efficient. Notably, SPFormer exceeds compared state-of-the-art methods by 4.3\% on ScanNetv2 hidden test set in terms of mAP and keeps fast inference speed (247ms per frame) simultaneously. Code is available at \url{https://github.com/sunjiahao1999/SPFormer}.
\end{abstract}

\section{Introduction}
\label{sec:intro}
3D scene understanding regards as a fundamental ingredient for many applications, including augmented/virtual reality \cite{ar}, autonomous driving \cite{ad}, and robotics navigation \cite{rn}. Generally, instance segmentation is a challenging task in 3D scene understanding, which aims to not only detect instances on sparse point clouds but also give a clear mask for each instance.

\begin{figure}[t]
\centering
\subfigure[Input Point Cloud]{
\includegraphics[width=0.36\linewidth]{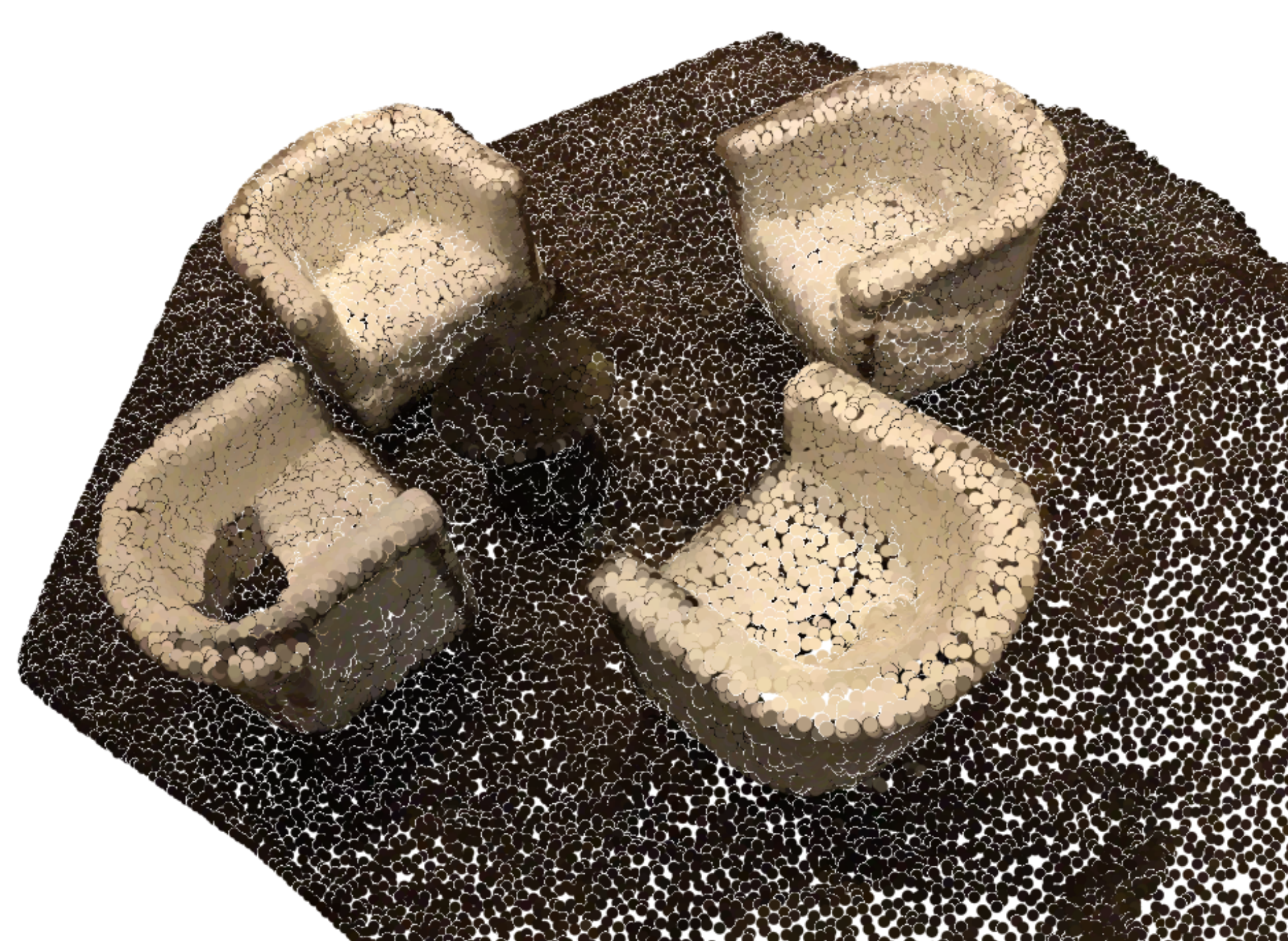}
}
\quad
\subfigure[Proposal-based]{
\label{fig:proposal-based}
\includegraphics[width=0.36\linewidth]{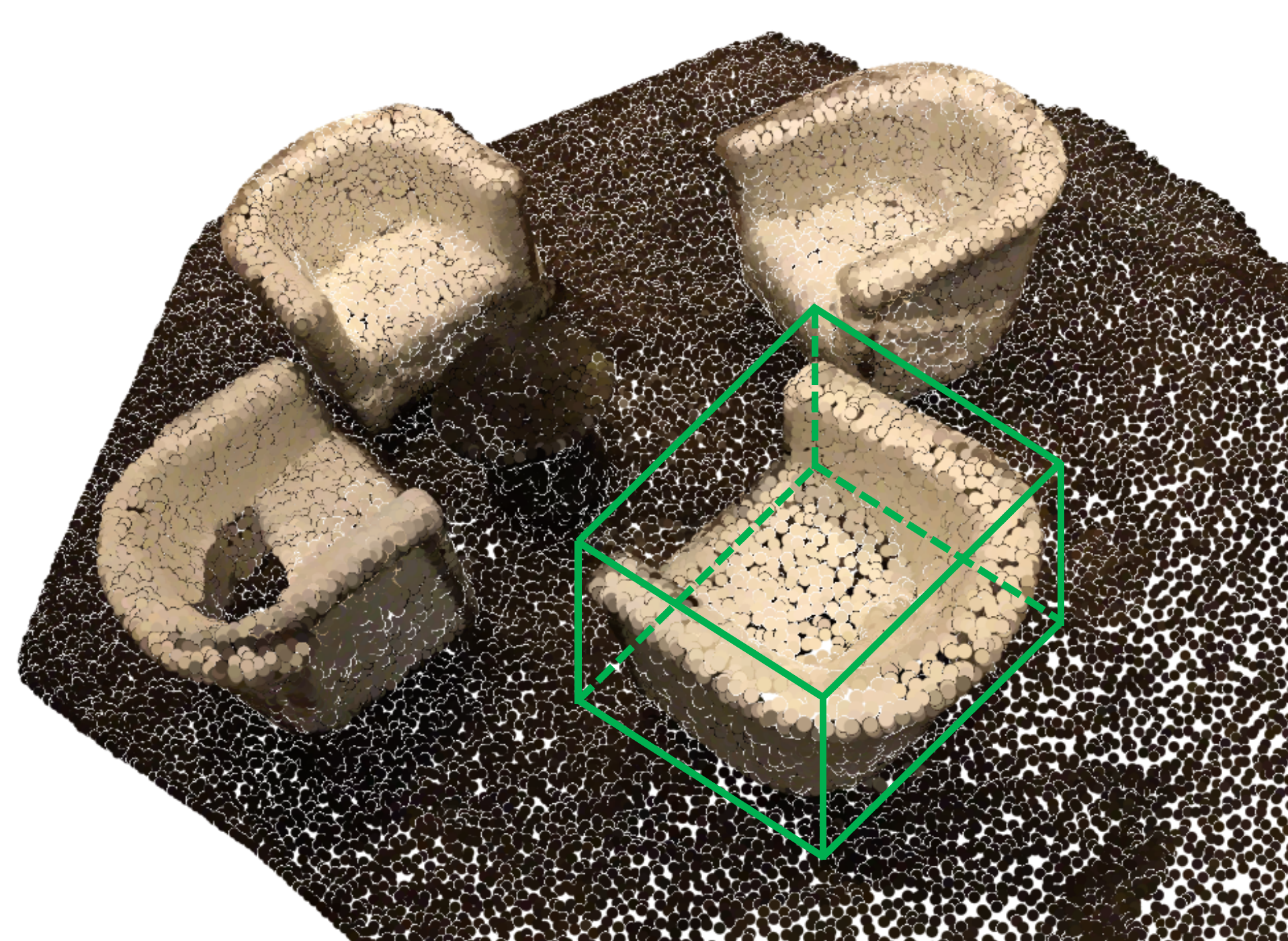}
}

\subfigure[Grouping-based]{
\label{fig:grouping-based}
\includegraphics[width=0.36\linewidth]{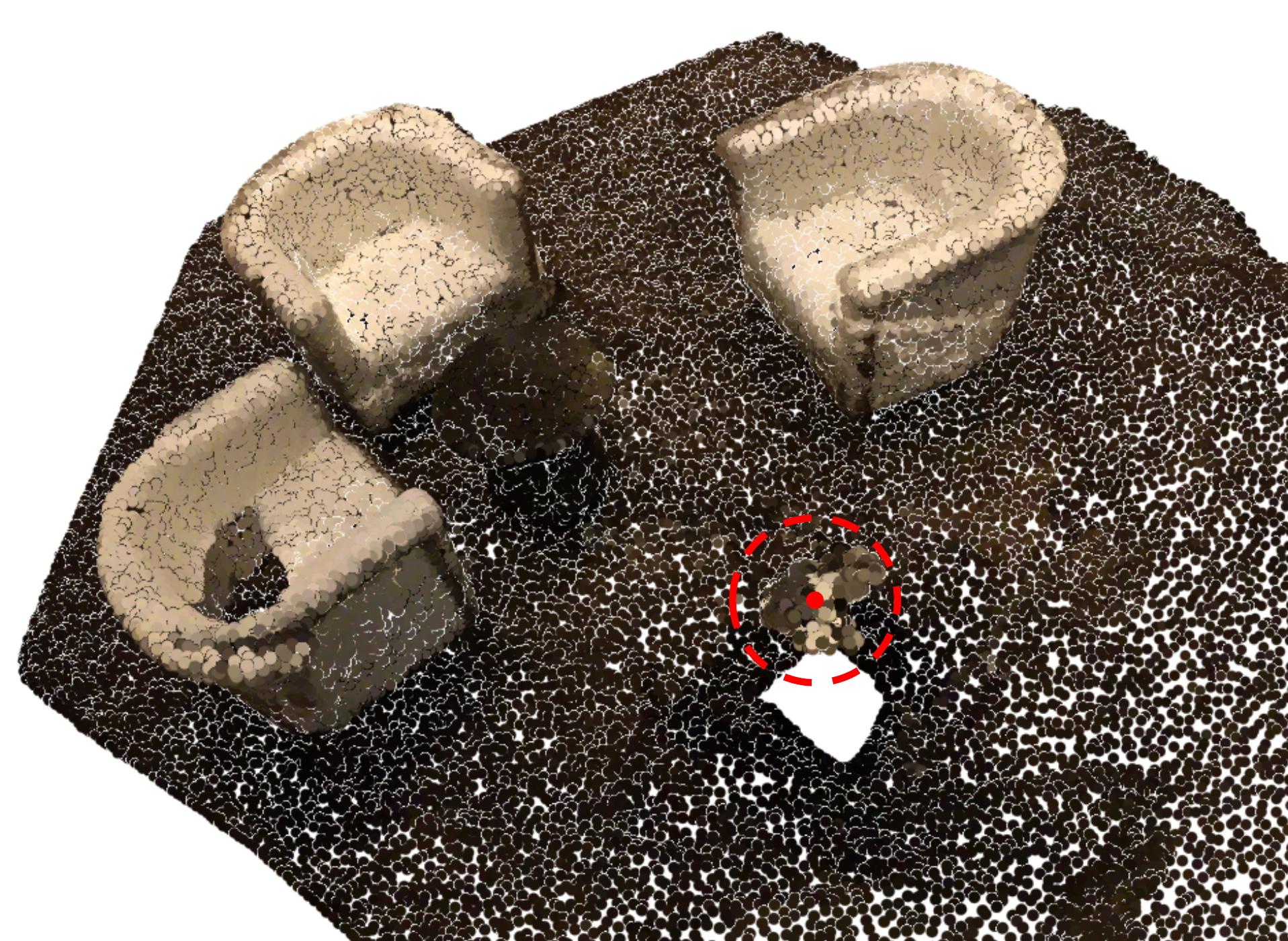}
}
\quad
\subfigure[Ours]{
\label{fig:ours}
\includegraphics[width=0.36\linewidth]{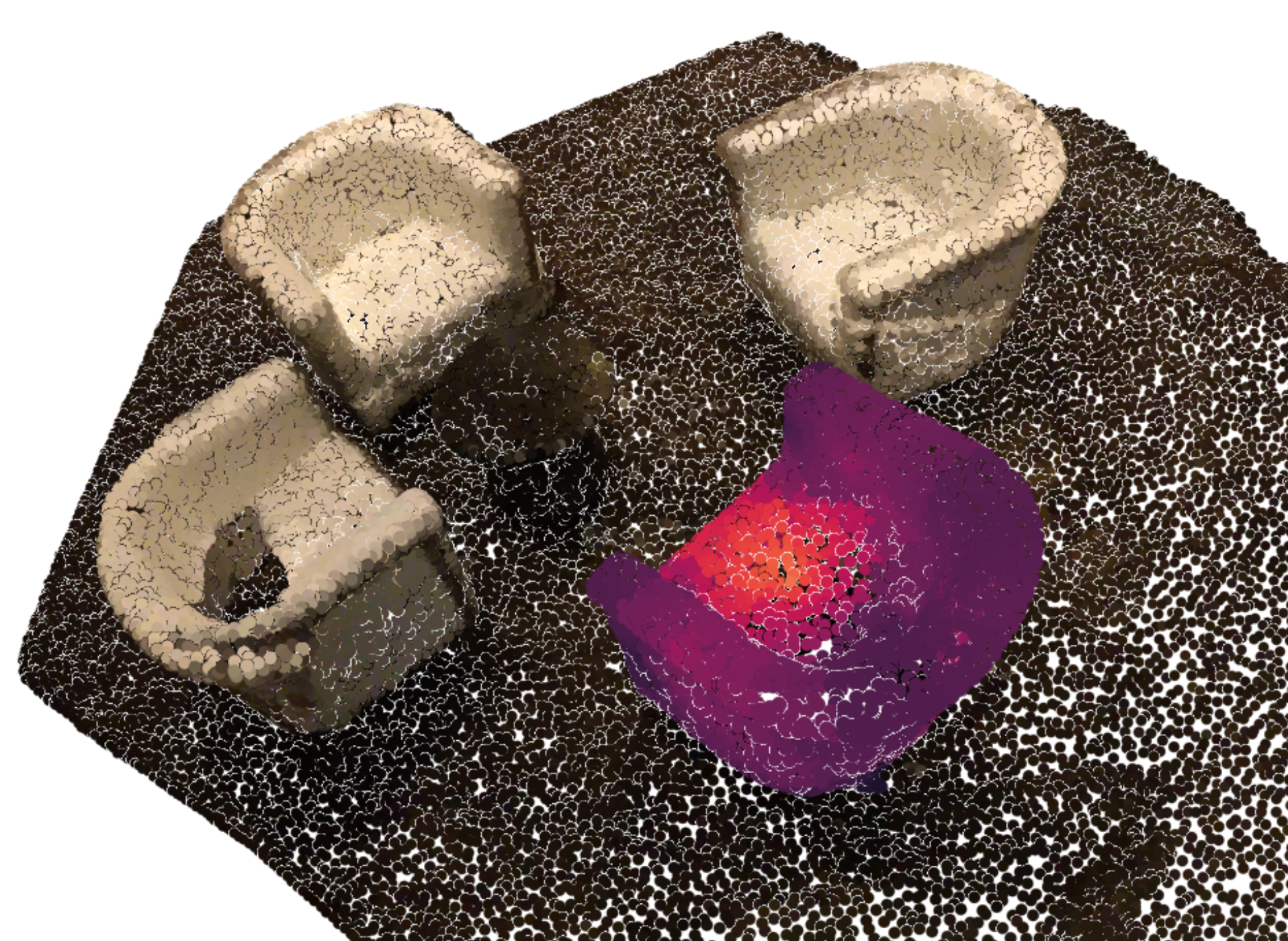}
}
\caption{\textbf{Key process of different methods}. (a) is an input point cloud. (b) Proposal-based methods detect objects first. (c) Grouping-based methods offset points to their own instance center and group points. (d) Our method highlights the region of interest by superpoint cross-attention.}
\end{figure}

Existing state-of-the-art methods can be divided into proposal-based \cite{3dbonet, gicn} and grouping-based \cite{pointgroup, hais, sstnet, softgroup}. Proposal-based methods consider 3D instance segmentation as a top-down pipeline. They firstly generate region proposals (i.e. bounding box), as shown in Fig. \ref{fig:proposal-based}, and then predict instance masks in the proposed region. These methods are encouraged by the big success of Mask-RCNN \cite{mask_rcnn} on 2D instance segmentation fields. However, these methods struggle on point clouds due to domain gaps. In 3D fields, bounding box has more degree of freedom (DoF) increasing the difficulty of fitting. Moreover, points usually only exist on parts of object surface, which causes object geometric centers to be not detectable. Besides, low-quality region proposals affect box-based bipartite matching \cite{3dbonet} and further degrade model performance. 

On the contrary, grouping-based methods adopt a bottom-up pipeline. They learn point-wise semantic labels and instance center offsets. Then they use the offsetted points and semantic predictions to aggregate into instances, as shown in Fig. \ref{fig:grouping-based}. Over the past two years, grouping-based methods have achieved great improvements in 3D instance segmentation task \cite{sstnet, softgroup}. However, there also are several shortcomings: (1) grouping-based methods depend on their semantic segmentation results, which might lead to wrong predictions. Propagating these wrong predictions to subsequent processing suppresses the performance of network. (2) These methods need an intermediate aggregation step increasing training and inference time. The aggregation step is independent of network training and lack of supervision, which needs an additional refinement module.

With the discussion above, we naturally think about a hyper framework that can avoid drawbacks and take benefits from two types of methods simultaneously. In this paper, we proposed a novel end-to-end two-stage 3D instance segmentation method based on Superpoint Transformer, named as \textbf{SPFormer}. SPFormer groups bottom-up potential features from point clouds into superpoints and proposes instances by query vectors as a top-down pipeline. 

In the bottom-up grouping stage, a sparse 3D U-net is utilized to extract bottom-up point-wise features. A simple superpoint pooling layer is presented to group potential point-wise features into superpoints. Superpoints \cite{superpoint} can leverage the geometric regularities to represent homogeneous neighboring points. In contrast to previous method \cite{sstnet}, our superpoint features are potential, which avoid supervising the features through non-straightforward semantic and central distance labels. We consider superpoints as a potential mid-level representation of 3D scenes and directly use instance labels to train the whole network. In the top-down proposal stage, a novel query decoder with transformers is proposed. We utilize learnable query vectors to propose instance prediction from potential superpoint features as a top-down pipeline. The learnable query vector can capture instance information through superpoint cross-attention mechanism. Fig. \ref{fig:ours} illustrates this process that the redder the part of the chair is, the more attention of query vector pays. With the query vectors carrying instance information and superpoint features, query decoder directly generates instance class, score, and mask predictions. Finally, through bipartite matching based on superpoint masks, SPFormer can implement end-to-end training without time-consuming aggregation step. Besides, SPFormer is free of post-processing like non-maximum suppression (NMS), which further accelerates the speed of network.

SPFormer achieves state-of-the-art on both ScanNetv2 and S3DIS benchmarks. Especially, SPFormer exceeds compared state-of-the-art methods by qualitative and quantitative measures, and inference speed, simultaneously. SPFormer with a novel pipeline can be served as a general framework for 3D instance segmentation. In summary, our contributions are listed as follows:

\begin{itemize}
    \item We propose a novel end-to-end two-stage method named SPFormer that represents 3D sence with potential superpopint features without relying on the results of object detection or semantic segmentation.
    \item We design a query decoder with transformers where learnable query vectors can capture instance information by superpoint cross-attention. With query vectors, query deocoder can directly generate instance predictions.
    \item Through bipartite matching based on superpoint masks, SPFormer can implement the network training without time-consuming intermediate aggregation step and be free of complex post-processing during inference.
\end{itemize}

\begin{figure*}[htbp]
    \centering
    \includegraphics[width=0.95\textwidth]{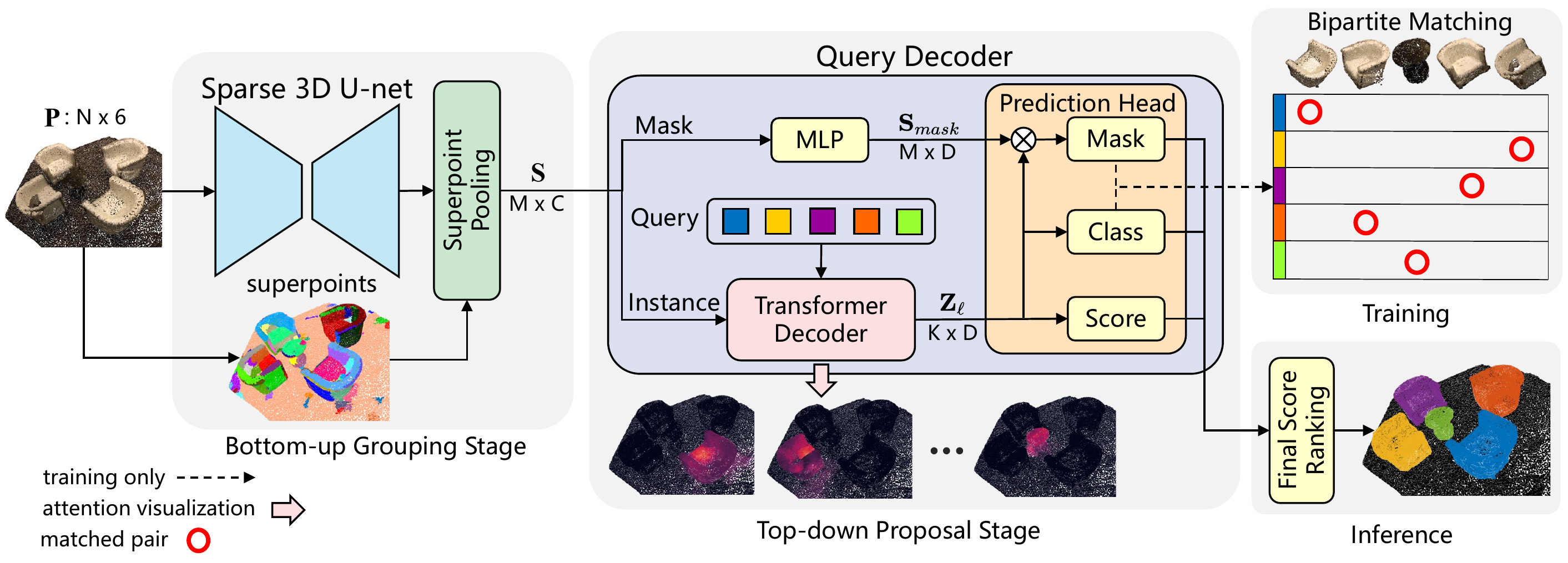}
    \caption{The overall architecture of SPFormer, which contains two stages. In the bottom-up grouping stage, sparse 3D U-net extracts point-wise features from input point cloud $\mathbf P $, and then superpoint pooling layer groups homogeneous neighboring points into superpoint features $\mathbf S $. In the top-down proposal stage, the query decoder is divided into two branches. The instance branch obtains query vector features $\mathbf{Z}_\ell $ by transformer decoder. The mask branch extracts mask-aware features $\mathbf S_{mask} $. Finally, a prediction head generates instance predictions and feeds them into bipartite matching or ranking during training/inference. }
    \label{fig:superpointformer}
\end{figure*}

\section{Related Work}
\label{sec:relwork}

\paragraph{Proposal-based Methods.} Proposal-based methods take a top-down pipeline for instance segmentation. Previous methods \cite{gspn,3d-sis,panopticfusion} focus on fusing 2D image features with point cloud features into a volumetric grid and generate region proposals from the grid. 3D-BoNet \cite{3dbonet} uses PointNet++ \cite{pointnet,pointnet++} extracting features from point clouds and treats 3D bounding box generation task as an optimal assignment problem. GICN \cite{gicn} predicts Gaussian heatmap to select instance center candidates and produces instance masks within the proposed bounding boxes. 3D-MPA \cite{3d-mpa} samples predicted centroids and cluster points near the centroids to form final instance masks. Most proposal-based methods are based on 3D bounding boxes. However, low-quality bounding boxes predictions will affect the performance of the instance segmentation model.

\paragraph{Grouping-based Methods.} Grouping-based methods regard 3D instance segmentation as a bottom-up pipeline. MTML \cite{mtml} utilizes a multi-task strategy to learn feature embedding. PointGroup \cite{pointgroup} aggregates points from original and center-shifted point clouds and designs ScoreNet for evaluating the quality of aggregation. PE\cite{pe} introduces a novel probabilistic embedding space. Dyco3D\cite{dyco3d} introduces dynamic convolution kernels. HAIS \cite{hais} extends PointGroup with a hierarchical aggregation and filters noisy points within instance prediction. SSTNet \cite{sstnet} constructs a semantic superpoint tree and gains instance prediction by splitting non-similar nodes. SoftGroup \cite{softgroup} uses a lower threshold for clustering to address the wrong semantic hard prediction and refines instances with a tiny 3D U-net. Although, grouping-based methods may have a top-down refinement module, they still inevitably rely on intermediate aggregation step.

\paragraph{2D Instance Segmentation with Transformer.} Recently, transformer \cite{transformer} is introduced in image classification \cite{vit, deit, swin}, object detection \cite{detr, up-detr} and segmentation \cite{maskformer,mask2former,sotr}. There are also some instance segmentation methods \cite{queryinst, sparseinst} inspired by transformer. Mask2Former \cite{mask2former}  successfully applies transformer to build a universal network for 2D image semantic, instance, and panoptic segmentation.

Inspired by the success of transformer for 2D segmentation tasks, we are motivated to introduce transformer for 3D instance segmentation. However, transformer cannot be naively applied on the output of sparse convolution backbone, because it will introduce highly computational overhead because of the complexity of attention mechanism. In this paper, we will design a novel query decoder for 3D instance segmentation and employ superpoints to build a bridge between the backbone and query decoder.

\section{Method}
\label{sec:method}
The architecture of the proposed SPFormer is illustrated in Fig. \ref{fig:superpointformer}. Firstly, a sparse 3D U-net is utilized to extract bottom-up point-wise features. A simple superpoint pooling layer is presented to group potential point-wise features into superpoints. Secondly, a novel query decoder with transformers is proposed, where learnable query vectors can capture instance information by superpoint cross-attention. Finally, through bipartite matching based on superpoint masks, SPFormer can implement end-to-end training without time-consuming aggregation step. 

\subsection{Backbone and Superpoints}
\label{sec:backbone}
\paragraph{Sparse 3D U-net.} Assuming that the input point cloud has $N$ points, the input can be expressed as $\mathbf P \in \mathbb{R} {^{N \times 6}}$. Each Point has colors $r$,$g$,$b$ and coordinates $x$,$y$,$z$. Following previous implementation \cite{spconv-unet}, we voxelize point cloud for regular input and use a U-net style backbone composed of submanifold sparse convolution (SSC) or sparse convolution (SC) to extract point-wise features $\mathbf P' \in \mathbb{R} {^{N \times C}}$. We give the sparse 3D U-net specifics in the supplementary material. Different from the common grouping-based methods, our method does not add an additional semantic branch and offset branch. 

\paragraph{Superpoint pooling layer.} To build an end-to-end framework, we directly feed point-wise features $\mathbf P' \in \mathbb{R} {^{N \times C}}$ into superpoint pooling layer based on pre-computed superpoints \cite{superpoint}. Superpoint pooling layer simply obtains superpoint features $\mathbf S \in \mathbb{R} {^{M \times C}}$ via average pooling over those point-wise ones inside each of superpoints. Without loss of generality, we suppose that there are $M$ superpoints computed from the input point cloud. Notably, superpoint pooling layer reliably downsample input point cloud to hundreds of superpoints, which significantly reduces the computational overhead of subsequent processing and optimizes the representation capability of the entire network.

\subsection{Query Decoder}
\label{sec:query decoder}
Query decoder consists of instance branch and mask branch. In the mask branch, a simple Multi-Layer Perceptron (MLP) aims to extract the mask-aware features $\mathbf{S}_{mask} \in \mathbb{R}{^{M \times D}}$. The instance branch is composed of a series of transformer decoder layers. They decode learnable query vectors via superpoint cross-attention. Assume there are $K$ learnable query vectors. We predefine the features of query vectors from each transformer decoder layer as $\mathbf{Z}_\ell \in \mathbb{R}{^{K \times D}}$. $D$ is embedding dimension and $\ell=1,2,3 ...$ is layer index. 

\begin{figure}[t]
    \centering
    \includegraphics[width=0.7\linewidth]{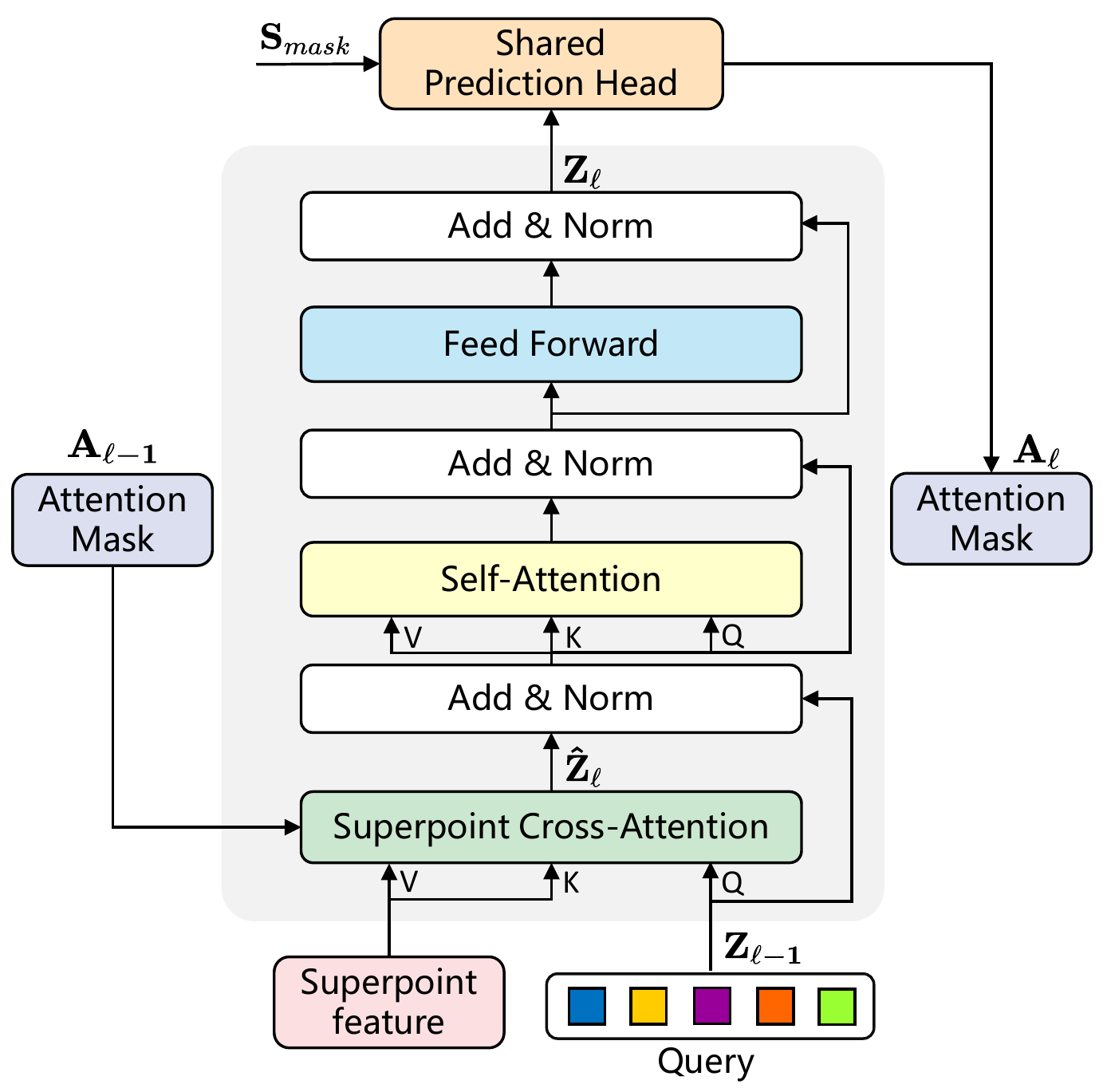}
    \caption{The architecture of transformer decoder layer and iterative prediction process. Here omits the branch that feeds the output query features $\mathbf{Z_\ell}$ to the next layer for readability.}
    \label{fig:decoder}
\end{figure}

\paragraph{Superpoint Cross-Attention.} Considering the disorder and quantity uncertainty of superpoint, transformer structure is introduced to handle variable length input. The potential feature of superpoints and the learnable query vectors are used as the input of the transformer decoder. The detailed architecture of our modified transformer decoder layer is depicted in Fig. \ref{fig:decoder}. Inspired by \cite{mask2former}, query vectors are initialized randomly before training, and the instance information of each point cloud can only be obtained through superpoint cross-attention, therefore, our transformer decoder layer exchanges the order of self-attention layer and cross-attention layer compared with the standard one\cite{transformer}. In addition, because the input is the potential features of superpoints, we empirically remove position embedding.

With superpoint features after linear projection $\mathbf{S'} \in \mathbb{R}{^{M \times D}}$, query vectors from former layer $\mathbf{Z_{\ell-1}}$ capture context information via superpoint cross-attention mechanism, which can be formulated as:
\begin{equation}
\label{equ:Zl}
    \mathbf{\hat{Z}_{\ell} }=softmax(\frac{\mathbf{Q}\mathbf{K}^T}{\sqrt[]{D}} +\mathbf{A}_{\ell-1})\mathbf{V},
\end{equation}
where $\mathbf{\hat{Z}_{\ell}} \in \mathbb{R}{^{K \times D}}$ is the output of superpiont cross-attention. $\mathbf{Q} = \psi_Q (\mathbf{Z_{\ell-1}}) \in \mathbb{R}{^{K \times D}}$ is the linear projection of input query vectors $\mathbf{Z_{\ell-1}}$ and $\mathbf{K}$, $\mathbf{V}$ is superpoint features $\mathbf{S'}$ with different linear projection $\psi_K(\cdot)$, $\psi_V(\cdot)$ respectively. $\mathbf{A_{\ell-1}} \in \mathbb{R}{^{K \times M}}$ is superpoint attention masks. Given the predicted superpoint masks $\mathbf{M}_{\ell-1}$ from the former prediction head, superpoint attention masks $\mathbf{A_{\ell-1}}$ filter superpoint with a threshold $\tau$, as 
\begin{equation}
\label{equ:A}
\mathbf{A_{\ell-1}}(i,j)=\left\{  
             \begin{array}{l} 
             \quad0\:\quad\quad\text{if $\:\:\mathbf{M_{\ell-1}}(i,j) \ge \tau$}\\
             -\infty\quad\quad \text{otherwise}
             \end{array}.
\right.
\end{equation}
$\mathbf{A_{\ell-1}}(i,j)$ indicates $i$-th query vector attending to $j$-th superpoint where $\mathbf{M_{\ell-1}}(i,j)$ is higher than $\tau$. Empirically, we set $\tau$ to 0.5. With transformer decoder layer stacking, superpoint attention masks $\mathbf{A_{\ell-1}}$ adaptively constrain cross-attention within the foreground instance.

\paragraph{Shared Prediction Head.} With query vectors $\mathbf{Z_\ell}$ from instance branch, we use two independent MLPs to predict the classification $\{p_i\in\mathbb{R}^{N_{class}+1} \}^K_{i=1}$ of each query vector and evaluate the quality of proposals with IoU-aware score $\{s_i\in [0, 1] \}^K_{i=1}$ respectively. Specifically, we append prediction with ``no instance" probability in addition to $N_{class}$ categories in order to assign ground truth to the proposals by bipartite matching and treat the other proposals as negative predictions. Moreover, the ranking of proposals profoundly affects instance segmentation results, while in practice most proposals will be regarded as background due to one-to-one matching style, which causes the misalignment of proposal quality ranking. Thus, We design a score branch that estimates the IoU of predicted superpoint masks and ground truth ones to compensate for the misalignment. 

Besides, given the mask-aware features $\mathbf{S}_{mask} \in \mathbb{R}{^{M \times D}}$ from mask branch, we directly multiply it by query vectors $\mathbf{Z_\ell}$ followed a sigmoid function to generate superpoint masks prediction $\mathbf{M_{\ell}} \in [0,1]^{K \times M} $.

\paragraph{Iterative Prediction.} Considering the slow convergence of transformer-based model \cite{detr}, we feed every transformer decoder layer output $\mathbf{Z}_\ell$ into the shared prediction head to generate proposals. Specially, we define $\mathbf{Z}_0$ to be the query vectors that have not captured instance information with 3D scene yet. we also feed $\mathbf{Z}_0$ into the shared prediction head, even if it is equivariant for any 3D scene. During training, we assign ground truth to all output from the shared prediction head with different layer input $\mathbf{Z_\ell}$. we find that it will improve the performance of model and query vectors feature will be updated layer by layer. We only use the output of the last prediction head for final instance proposals during inference, which can avoid the redundancy of proposals and accelerate inference speed. 

\subsection{Bipartite Matching and Loss Function}
\label{sec:bl}
With a fixed number of proposals, we formulate ground truth label assignment as an optimal assignment problem. Formally, we introduce a pairwise matching cost $\mathcal{C}_{ik}$ to evaluate the similarity of the $i$-th proposal and the $k$-th ground truth. $\mathcal{C}_{ik}$ is determined by classification probability and superpoint mask matching cost $\mathcal{C}^{mask}_{ik}$, as defined in Eq. (\ref{equ:C}).
\begin{equation}
\label{equ:C}
\mathcal{C}_{ik} = -\lambda_{cls}\cdot  p_{i,c_k}+\lambda_{mask}\cdot \mathcal{C}^{mask}_{ik},
\end{equation}
where $p_{i,c_k}$ indicates the probability for the category $c_k$ of $i$-th proposal and $\lambda_{cls}$, $\lambda_{mask}$ are corresponding coefficients of each term. In our experiments, we set $\lambda_{cls} = 0.5$, $\lambda_{mask}=1$. Superpoint mask matching cost $\mathcal{C}^{mask}_{ik}$ consists of binary cross-entropy (BCE) and dice loss with Laplace smoothing \cite{VNet}, as
\begin{equation}
\label{equ:Cmask}
\mathcal{C}^{mask}_{ik} = \text{BCE}(m_i,m^{gt}_k)+1-2\frac{ m_i\cdot m^{gt}_k+1}{\left | m_i \right | +\left | m^{gt}_k \right |+1},
\end{equation}
where $m_i$ and $m^{gt}_k$ are the superpoint mask of proposal and ground truth respectively. We assign a hard instance label to each superpoint depending on whether more than half of the points within the superpoint belong to the instance. With the matching cost $\mathcal{C}_{ik}$, we use Hungarian algorithm \cite{hungarian} to find the optimal matching between proposals and ground truth.

\begin{table*}[!htb] 
\centering
\resizebox{\textwidth}{!}{
\begin{tabular}{l|c|cccccccccccccccccc}
    \toprule
        \multirow{3}{*}{Method} & \multicolumn{1}{l|}{\multirow{3}{*}{mAP}} & \multirow{3}{*}{\rotatebox{90}{bath}} & \multirow{3}{*}{\rotatebox{90}{bed}} & \multirow{3}{*}{\rotatebox{90}{bkshf}} & \multirow{3}{*}{\rotatebox{90}{cabinet}} & \multirow{3}{*}{\rotatebox{90}{chair}} & \multirow{3}{*}{\rotatebox{90}{counter}} & \multirow{3}{*}{\rotatebox{90}{curtain}} & \multirow{3}{*}{\rotatebox{90}{desk}} & \multirow{3}{*}{\rotatebox{90}{door}} & \multirow{3}{*}{\rotatebox{90}{other}} & \multirow{3}{*}{\rotatebox{90}{picture}} & \multirow{3}{*}{\rotatebox{90}{fridge}} & \multirow{3}{*}{\rotatebox{90}{s. cur.}} & \multirow{3}{*}{\rotatebox{90}{sink}} & \multirow{3}{*}{\rotatebox{90}{sofa}} & \multirow{3}{*}{\rotatebox{90}{table}} & \multirow{3}{*}{\rotatebox{90}{toilet}} & \multirow{3}{*}{\rotatebox{90}{wind.}} \\ 
        & \multicolumn{1}{l|}{} & \multicolumn{1}{l}{} & \multicolumn{1}{l}{} & \multicolumn{1}{l}{} & \multicolumn{1}{l}{} & \multicolumn{1}{l}{} & \multicolumn{1}{l}{} & \multicolumn{1}{l}{} & \multicolumn{1}{l}{} & \multicolumn{1}{l}{} & \multicolumn{1}{l}{} & \multicolumn{1}{l}{} & \multicolumn{1}{l}{} & \multicolumn{1}{l}{} & \multicolumn{1}{l}{} & \multicolumn{1}{l}{} & \multicolumn{1}{l}{} & \multicolumn{1}{l}{} & \multicolumn{1}{l}{}\\
        & \multicolumn{1}{l|}{} & \multicolumn{1}{l}{} & \multicolumn{1}{l}{} & \multicolumn{1}{l}{} & \multicolumn{1}{l}{} & \multicolumn{1}{l}{} & \multicolumn{1}{l}{} & \multicolumn{1}{l}{} & \multicolumn{1}{l}{} & \multicolumn{1}{l}{} & \multicolumn{1}{l}{} & \multicolumn{1}{l}{} & \multicolumn{1}{l}{} & \multicolumn{1}{l}{} & \multicolumn{1}{l}{} & \multicolumn{1}{l}{} & \multicolumn{1}{l}{} & \multicolumn{1}{l}{} & \multicolumn{1}{l}{}\\
    \midrule
    3D-BoNet & 25.3 & 51.9 & 32.4 & 25.1 & 13.7 & 34.5 & 3.1 & 41.9 & 6.9 & 16.2 & 13.1 & 5.2 & 20.2 & 33.8 & 14.7 & 30.1 & 30.3 & 65.1 & 17.8\\
    MTML& 28.2 & 57.7 & 38.0 & 18.2 & 10.7 & 43.0 & 0.1 & 42.2 & 5.7 & 17.9 & 16.2 & 7.0 & 22.9 & 51.1 & 16.1 & 49.1 & 31.3 & 65.0 & 16.2\\
    GICN & 34.1 & 58.0 & 37.1 & 34.4 & 19.8 & 46.9 & 5.2 & 56.4 & 9.3 & 21.2 & 21.2 & 12.7 & 34.7 & 53.7 & 20.6 & 52.5 & 32.9 & 72.9 & 24.1\\
    3D-MPA & 35.5 & 45.7 & 48.4 & 29.9 & 27.7 & 59.1 & 4.7 & 33.2 & 21.2 & 21.7 & 27.8 & 19.3 & 41.3 & 41.0 & 19.5 & 57.4 & 35.2 & 84.9 & 21.3\\
    Dyco3D & 39.5 & 64.2 & 51.8 & 44.7 & 25.9 & 66.6 & 5.0 & 25.1 & 16.6 & 23.1 & 36.2 & 23.2 & 33.1 & 53.5 & 22.9 & 58.7 & 43.8 & 85.0 & 31.7\\
    PE & 39.6 & 66.7 & 46.7 & 44.6 & 24.3 & 62.4 & 2.2 & 57.7 & 10.6 & 21.9 & 34.0 & 23.9 & 48.7 & 47.5 & 22.5 & 54.1 & 35.0 & 81.8 & 27.3\\
    PointGroup & 40.7 & 63.9 & 49.6 & 41.5 & 24.3 & 64.5 & 2.1 & 57.0 & 11.4 & 21.1 & 35.9 & 21.7 & 42.8 & 66.6 & 25.6 & 56.2 & 34.1 & 86.0 & 29.1\\
    HAIS & 45.7 & 70.4 & 56.1 & 45.7 & 36.4 & 67.3 & 4.6 & 54.7 & 19.4 & 30.8 & 42.6 & 28.8 & 45.4 & 71.1 & 26.2 & 56.3 & 43.4 & 88.9 & 34.4\\
    OccuSeg & 48.6 & \textbf{80.2} & 53.6 & 42.8 & 36.9 & 70.2 & 20.5 & 33.1 & 30.1 & 37.9 & 47.4 & 32.7 & 43.7 & \textbf{86.2} & 48.5 & 60.1 & 39.4 & 84.6 & 27.3\\
    SoftGroup & 50.4 & 66.7 & 57.9 & 37.2 & 38.1 & 69.4 & 7.2 & \textbf{67.7} & 30.3 & 38.7 & \textbf{53.1} & 31.9 & \textbf{58.2} & 75.4 & 31.8 & 64.3 & 49.2 & 90.7 & \textbf{38.8}\\
    SSTNet & 50.6 & 73.8 & 54.9 & \textbf{49.7} & 31.6 & 69.3 & 17.8 & 37.7 & 19.8 & 33.0 & 46.3 & \textbf{57.6} & 51.5 & 85.7 & \textbf{49.4} & 63.7 & 45.7 & \textbf{94.3} & 29.0\\ \hline
    \textbf{SPFormer} & \textbf{54.9} & 74.5 & \textbf{64.0} & 48.4 & \textbf{39.5} & \textbf{73.9} & \textbf{31.1} & 56.6 & \textbf{33.5} & \textbf{46.8} & 49.2 & 55.5 & 47.8 & 74.7 & 43.6 & \textbf{71.2} & \textbf{54.0} & 89.3 & 34.3\\
    \bottomrule
\end{tabular}}
\caption{3D instance segmentation results on ScanNetv2 hidden test set. Reported results are obtained from the ScanNet benchmark testing server on 11/07/2022.}
\label{tab:scannet_test}
\end{table*}

After assignment, we treat the proposals that are not assigned to ground truth as "no instance" class and compute the classification cross-entropy loss $\mathcal{L}_{cls}$ for every proposal. Then we compute the superpoint mask loss which consists of binary cross-entropy loss $\mathcal{L}_{bce}$ and dice loss $\mathcal{L}_{dice}$ for each proposal ground truth pair. In addition, we add the following L2 loss $\mathcal{L}_{s}$ for the score branch:
\begin{equation}
\label{equ:Ls}
\mathcal{L}_{s} = \frac{1}{\sum_{k=1}^{N_{gt}} \mathds{1}_{\{iou_k\}}} \sum_{k=1}^{N_{gt}}\mathds{1}_{\{iou_k\}}\left \| s_k-iou_k \right \|_2,
\end{equation}
where $\{s_k\}^{N_{gt}}_{k=1}$ is the set of score predictions that are assigned to $N_{gt}$ ground truth. $\mathds{1}_{\{iou_k\}}$ indicates whether the IoU between proposal mask prediction and assigned ground truth is higher than 50\%. We only use high-quality proposals for supervision \cite{ms_rcnn}. Finally, to build an end-to-end training, we adopt multi-task loss $\mathcal{L}$, as
\begin{equation}
\label{equ:L}
\mathcal{L}=\beta_{cls}\cdot \mathcal{L}_{cls}+\beta_s\cdot \mathcal{L}_{s}+\beta_{mask}\cdot (\mathcal{L}_{bce}+\mathcal{L}_{dice}),
\end{equation}
where $\beta_{cls}$, $\beta_{s}$, $\beta_{mask}$ are corresponding coefficients of each
term. Empirically, we set $\beta_{cls} = \beta_{s} = 0.5$, $\beta_{mask}=1$.

\subsection{Inference}
\label{sec:inference}
During inference, given an input point cloud, SPFormer directly predicts $K$ instances with classification $\{p_i\}$, IoU-aware score $\{s_i\}$ and corresponding superpoint masks. We additionally obtain a mask score $\{ms_i \in [0, 1]\}^K$ by averaging superpoints probability higher than 0.5 in each superpoint mask. The final score for sorting $\widetilde{s_i} = \sqrt[3]{p_i\cdot s_i \cdot ms_i}$. SPFormer is free of non-maximum suppression in post-processing, which ensures its fast inference speed.

\section{Experiments}
\label{sec:exps}

\paragraph{Datasets.} Experiments are conducted on ScanNetv2 \cite{scannet} and S3DIS \cite{s3dis} datasets. ScanNetv2 has a total of 1613 indoor scenes, of which 1201 are used for training, 312 for validation, and 100 for testing. It contains 18 categories of object instances. We submit the final prediction of our method to its hidden test set and the ablation studies are conducted on its validation set. S3DIS has 6 large-scale areas with 272 scenes in total. It has 13 categories for instance segmentation task. We follow two common settings for evaluation: testing on Area 5 and 6-fold cross-validation.


\paragraph{Evaluation Metrics.} Task-mean average precision (mAP) is utilized as the common evaluation metric for instance segmentation, which averages the scores with IoU thresholds set from 50\% to 95\%, with a step size of 5\%. Specifically, $\text{AP}_{50}$ and $\text{AP}_{25}$ denote the scores with IoU thresholds of 50\% and 25\%, respectively. We report mAP, $\text{AP}_{50}$ and $\text{AP}_{25}$ on ScanNetv2 dataset and we addtionally report mean precision (mPrec), and mean recall (mRec) on S3DIS dataset.

\subsection{Benchmark Results}
\paragraph{ScanNetv2.} SPFormer is compared with existing state-of-the-art methods on the hidden test set, as shown in Table \ref{tab:scannet_test}. SPFormer accomplishes the highest mAP score of 54.9\%, outperforming the previous best result by 4.3\%. For the specific 18 categories, our model achieves the highest AP scores on 8 of them. Especially, SPFormer surpasses the previous best AP score by more than 10\% in the \emph{counter} category,  where past methods are always hard to achieve a satisfactory score. 

We also evaluate SPFormer on ScanNetv2 validation set, as shown in Table \ref{tab:scannet_val}. SPFormer outperforms all state-of-the-art methods by a large margin. Compared to the second-best results, our method improves 6.9\%, 6.3\%, 4.0\% in terms of mAP, $\text{AP}_{50}$ and $\text{AP}_{25}$, respectively.

\paragraph{S3DIS.} We evaluate SPFormer on S3DIS using Area 5 and 6-fold cross-validation, respectively. As shown in Table \ref{tab:s3dis}, SPFormer achieves the-state-of-art results in terms of $AP_{50}$. Following the protocols used in previous methods, we additionally report mPrec and mRec. Our method also achieves competitive results in mPrec/mRec metrics. The results on S3DIS confirm the generalization ability of SPFormer.

\paragraph{Runtime Analysis.} 
We test the runtime per scene of different methods on ScanNetv2 validation set, as shown in Table \ref{tab:inf_speed}. For a fair comparison, the SSC and SC layers in all the above methods are implemented by spconv v2.1 \cite{spconv2}. We report in detail the running time of the components of each method (the last part of each model contains their own post-processing). Since our SPFormer and \cite{sstnet} is based on superpoints, here we add superpoints extraction (s.p. extraction) runtime to test the inference speed from raw input point clouds. However, superpoints can pre-compute in training stage, which can significantly reduce the model training time. Even with superpoints extraction, SPFormer is still the fastest method compared to the existing ones. 

\begin{table}[t]
    \centering
    \begin{tabular}{c|ccc}
        \toprule
        Method & mAP & $\text{AP}_{50}$ & $\text{AP}_{25}$ \\
        \midrule
        PE & 33.0 & 57.1 & 73.8 \\ 
        PointGroup &35.2 & 57.1 & 71.4 \\
        3D-MPA & 35.3 & 59.1 & 72.4 \\
        Dyco3D & 35.4 & 57.6 & 72.9 \\
        HAIS & 44.1 & 64.4 & 75.7 \\
        SoftGroup & 46.0 & 67.6 & 78.9 \\
        SSTNet & 49.4 & 64.3 & 74.0 \\
        \textbf{SPFormer} & \textbf{56.3} & \textbf{73.9} & \textbf{82.9} \\
        \bottomrule
    \end{tabular}
    \caption{3D instance segmentation results on ScanNetv2 validation set.}
    \label{tab:scannet_val}
\end{table}

\begin{table}[t]
    \centering
    \begin{tabular}{c|ccc}
        \toprule
        Method & $\text{AP}_{50}$ & $\text{mPrec}$ & $\text{mRec}$ \\
        \midrule
        PointGroup & 57.8 & 55.3 & 42.4 \\
        Dyco3D & - & 64.3 & 64.2 \\
        SSTNet & 59.3 & 65.5 & 64.2 \\
        HAIS & - & 71.1 & 65.0 \\
        SoftGroup & 66.1 & \textbf{73.6} & 66.6 \\
        \textbf{SPFormer} & \textbf{66.8} & 72.8 & \textbf{67.1} \\ \hline
        3D-BoNet\dag & - & 65.6 & 47.7 \\
        GICN\dag & - & 68.5 & 50.8 \\
        PointGroup\dag & 64.0 & 69.6 & 69.2 \\
        SSTNet\dag & 67.8 & 73.5 & \textbf{73.4} \\
        HAIS\dag & - & 73.2 & 69.4 \\
        SoftGroup\dag & 68.9 & \textbf{75.3} & 69.8 \\
        \textbf{SPFormer}\dag & \textbf{69.2} & 74.0 & 71.1 \\
        \bottomrule
    \end{tabular}
    \caption{3D instance segmentation results on the S3DIS validation set. Methods marked without $\dag$ are evaluated on Area 5; methods marked with $\dag$ are evaluated on 6-fold cross-validation.}
    \label{tab:s3dis}
\end{table}

\begin{table}[t]
    \centering
    \resizebox{\linewidth}{!}{
    \begin{tabular}{clc}
        \toprule
        Method & Component time (ms) & Total (ms) \\
        \midrule
        \multirow{3}{*}{PointGroup} & \multicolumn{1}{l}{Backbone(GPU):48} &  \multirow{3}{*}{372} \\
        \multicolumn{1}{l}{} & \multicolumn{1}{l}{Grouping(GPU+CPU):218} & \multicolumn{1}{l}{} \\
        \multicolumn{1}{l}{} & \multicolumn{1}{l}{ScoreNet(GPU):106} & \multicolumn{1}{l}{} \\ \hline
        
        \multirow{3}{*}{HAIS} & \multicolumn{1}{l}{Backbone(GPU):50} &  \multirow{3}{*}{256} \\
        \multicolumn{1}{l}{} & \multicolumn{1}{l}{Hier. aggr.(GPU+CPU): 116} & \multicolumn{1}{l}{} \\
        \multicolumn{1}{l}{} & \multicolumn{1}{l}{Intra-inst refinement(GPU): 90} & \multicolumn{1}{l}{} \\ \hline
        
        \multirow{3}{*}{SoftGroup} & \multicolumn{1}{l}{Backbone(GPU):48} &  \multirow{3}{*}{266} \\
        \multicolumn{1}{l}{} & \multicolumn{1}{l}{Soft grouping(GPU+CPU):121} & \multicolumn{1}{l}{} \\
        \multicolumn{1}{l}{} & \multicolumn{1}{l}{Top-down refinement(GPU):97} & \multicolumn{1}{l}{} \\ \hline
        
        \multirow{4}{*}{SSTNet} & \multicolumn{1}{l}{S.p. extraction(CPU):179} & \multirow{4}{*}{419} \\
        \multicolumn{1}{l}{} & \multicolumn{1}{l}{Backbone(GPU):34} & \multicolumn{1}{l}{} \\
        \multicolumn{1}{l}{} & \multicolumn{1}{l}{Tree Network(GPU+CPU):148} & \multicolumn{1}{l}{} \\
        \multicolumn{1}{l}{} & \multicolumn{1}{l}{ScoreNet(GPU):58} & \multicolumn{1}{l}{} \\ \hline
        
        \multirow{4}{*}{\textbf{SPFormer}} & \multicolumn{1}{l}{S.p. extraction(CPU):179} & \multirow{4}{*}{\textbf{247}} \\
        \multicolumn{1}{l}{} & \multicolumn{1}{l}{Backbone(GPU):29} & \multicolumn{1}{l}{} \\
        \multicolumn{1}{l}{} & \multicolumn{1}{l}{S.p. pooling(GPU):18} & \multicolumn{1}{l}{} \\
        \multicolumn{1}{l}{} & \multicolumn{1}{l}{Query decoder(GPU):21} & \multicolumn{1}{l}{} \\
        \bottomrule
    \end{tabular}}
    \caption{Inference time per scan of different methods on ScanNetv2 validation set. The runtime is measured on the same RTX 3090 GPU.}
    \label{tab:inf_speed}
\end{table}

\subsection{Ablation Study}
\paragraph{Components Analysis.} Table \ref{tab:key component} shows the performance results when different components are omitted. Considering naively feeding the output of backbone into query decoder, we find that there is a huge drop in performance. Query vectors can not attend to several hundred thousand points due to the softmax process in cross-attention. We employ superpoints to build a bridge between backbone and query decoder, which significantly improves our method performance. Then we discuss the bipartite matching target. We compare matching by boxes with matching by masks. The detail of the implementation of matching by boxes is in the supplementary material. We find the performance of matching by mask exceeds box one by 6.4\% on mAP. 3D boxes have more DoF than 2D ones and object geometric centers are usually not detectable, which inevitably makes matching more difficult. Finally, we confirm IoU-aware score branch brings benefits to our method. It takes +1.3/1.5/0.4 improvements on mAP/$\text{AP}_{50}$/$\text{AP}_{25}$ respectively. The score branch mitigates the misalignment of proposal quality ranking.

\paragraph{The Architecture of Transformer.} The ablation analysis of the architecture of transformer is illustrated in Table \ref{tab:transformer component}. Considering the original transformer decoder layer \cite{transformer} without position encoding as baseline, iteratively predicting on each transformer layer by the shared prediction head can bring +1.5/1.8/1.8 improvement on mAP/$\text{AP}_{50}$/$\text{AP}_{25}$ respectively. Moreover, if we add superpoint attention masks, our method will further improve +3.8/2.4/1.3  performance. Superpoint attention masks allow SPFormer to only attend to the foreground from the former layer predictions. Due to the uncertainty of the number of superpoints in each scene, we only discuss whether use position encoding on query vectors. We add position encoding where query vectors are fed into every decoder layer. We observe that the position encoding can safely remove, probably due to the irregularity and diversity of the point clouds. At last, we swap the order of self-attention and cross-attention, for query vectors can gather context information immediately once they are fed into decoder layer, which makes the process more sensible and brings a little improvement.

\begin{figure*}[t]
    \centering
    \begin{minipage}[htbp]{0.19\textwidth}
    \centering
    \includegraphics[width=0.95\linewidth]{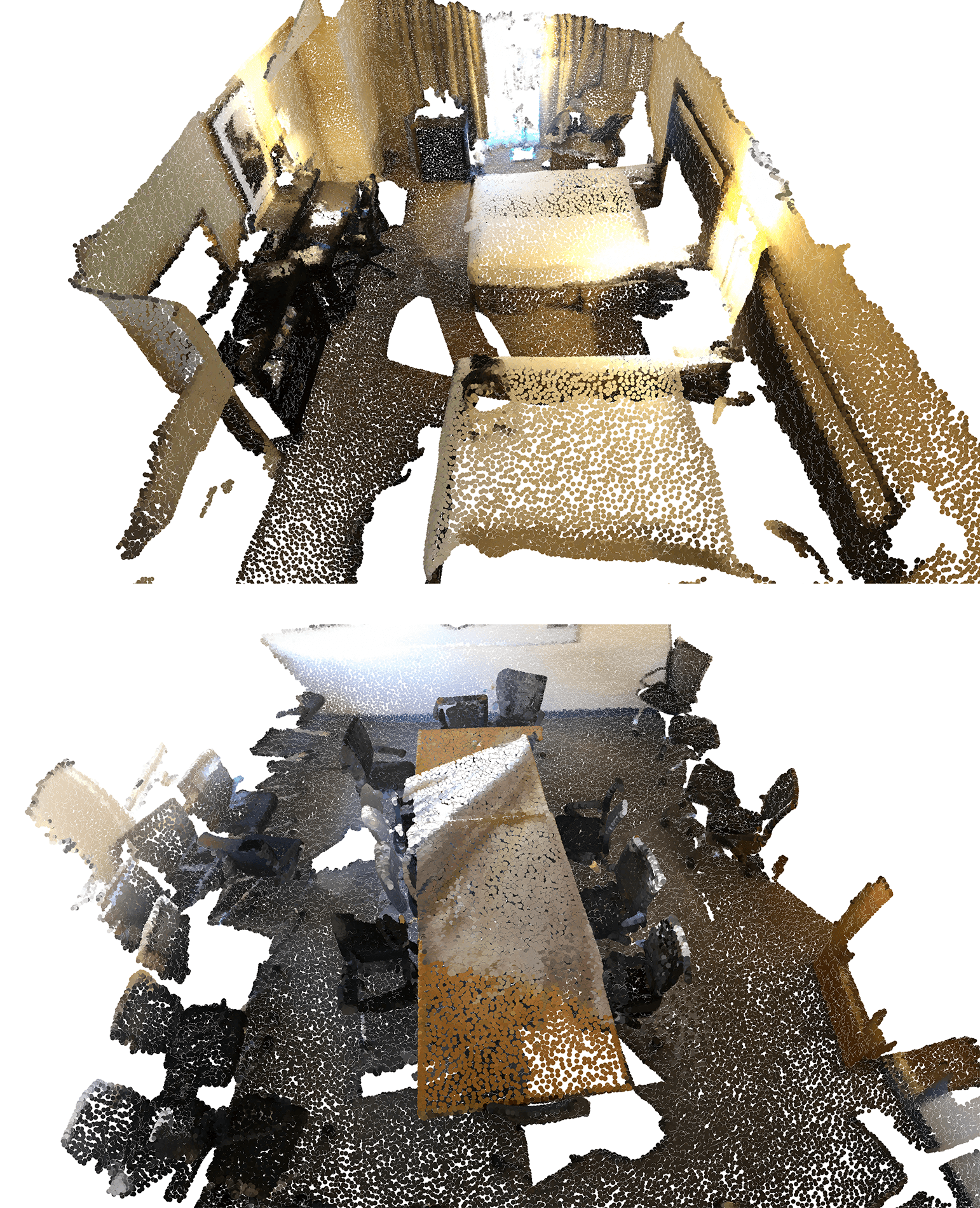}\\
    Input
    \end{minipage}%
    \begin{minipage}[htbp]{0.19\textwidth}
    \centering
    \includegraphics[width=0.95\linewidth]{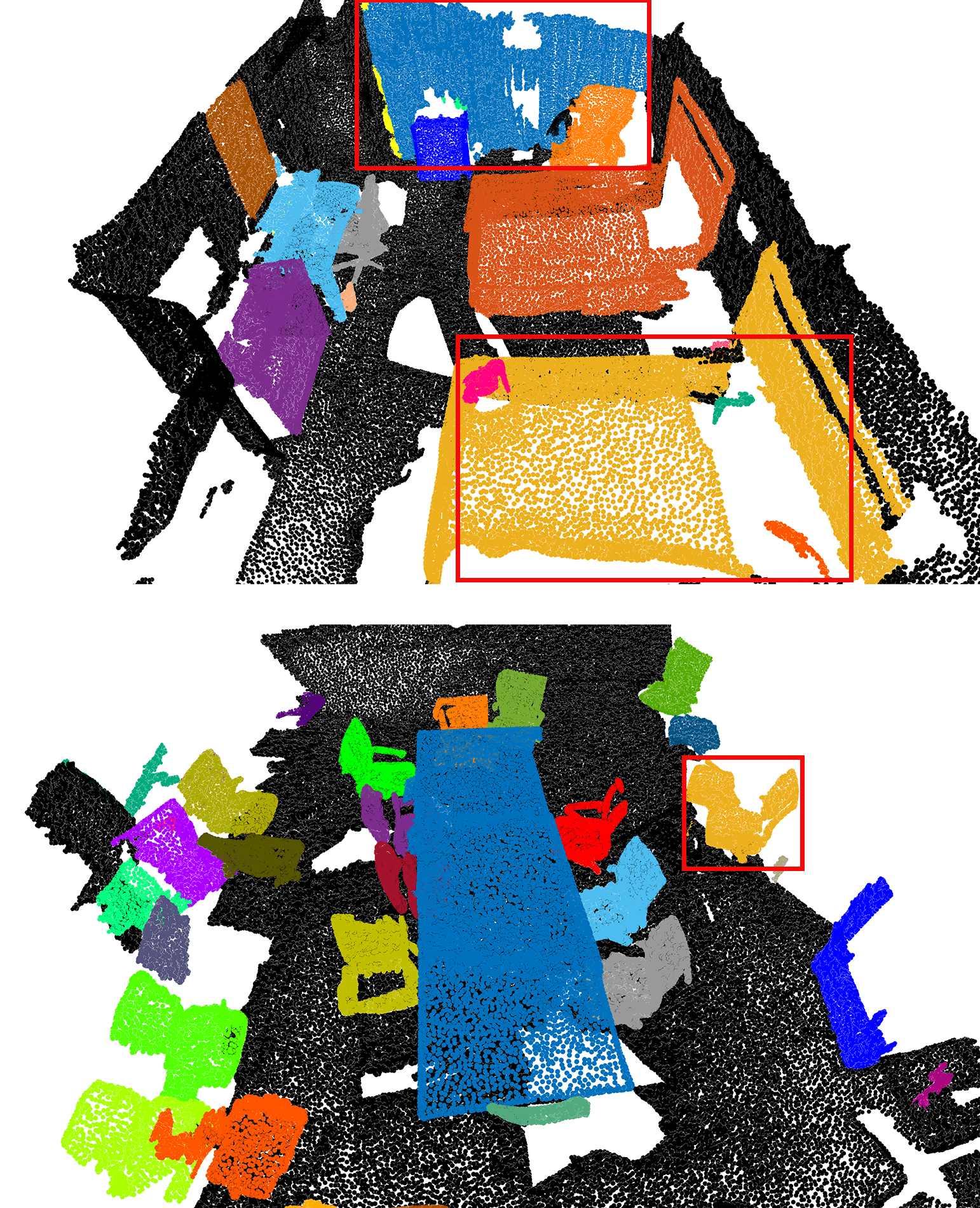}\\
    SSTNet
    \end{minipage}%
    \begin{minipage}[htbp]{0.19\textwidth}
    \centering
    \includegraphics[width=0.95\linewidth]{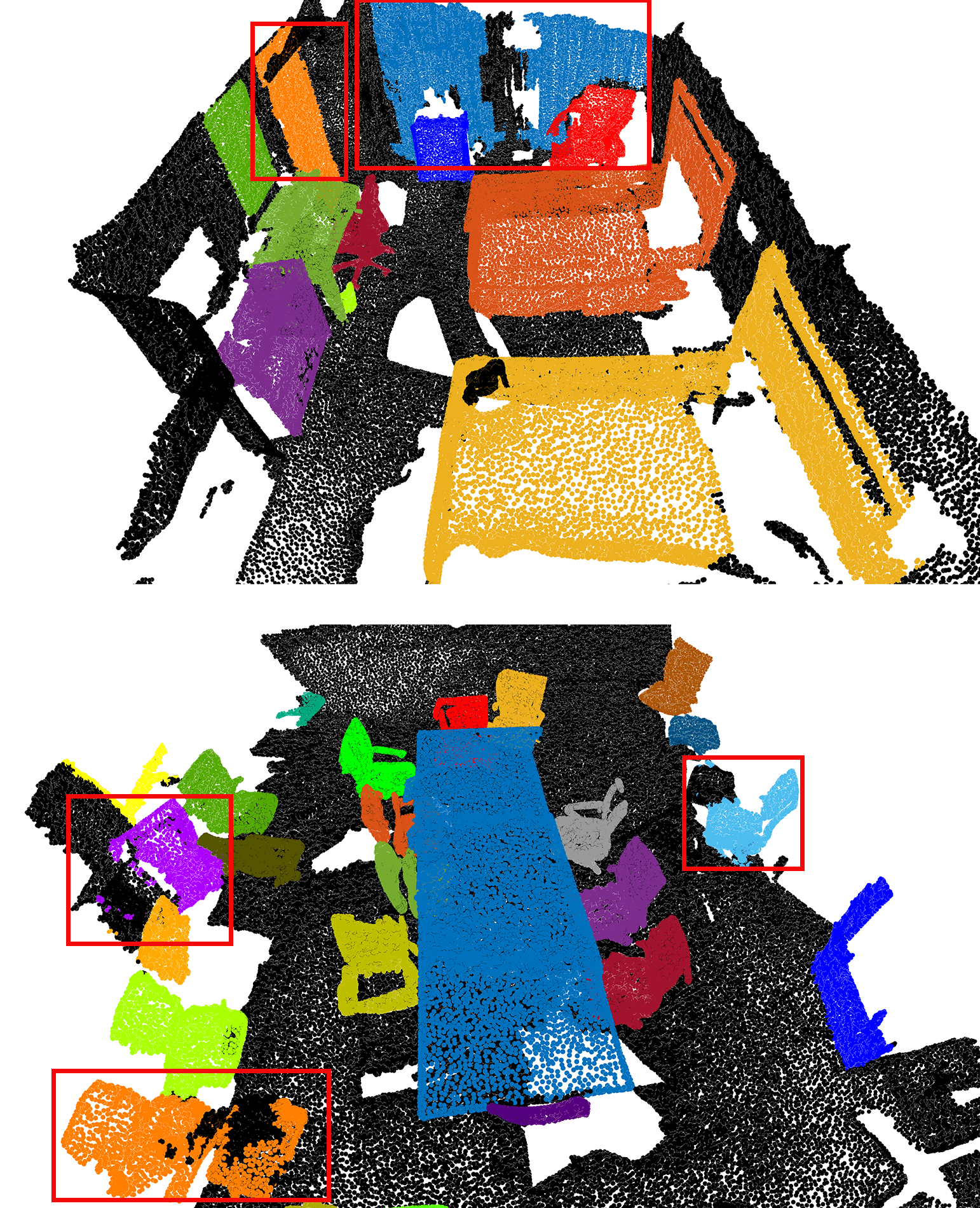}\\
    SoftGroup
    \end{minipage}%
    \centering
    \begin{minipage}[htbp]{0.19\textwidth}
    \centering
    \includegraphics[width=0.95\linewidth]{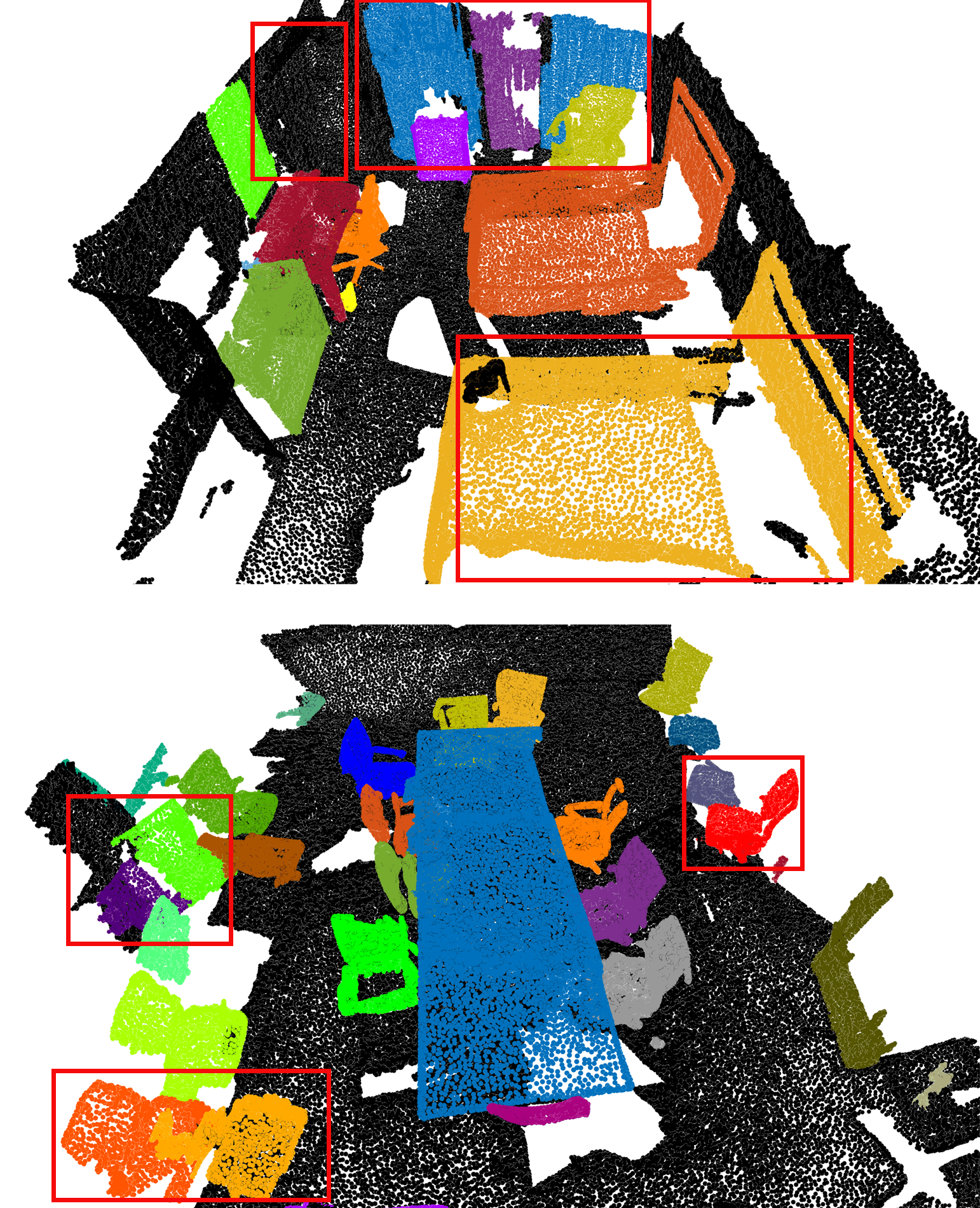}\\
    Ours
    \end{minipage}%
    \begin{minipage}[htbp]{0.19\textwidth}
    \centering
    \includegraphics[width=0.95\linewidth]{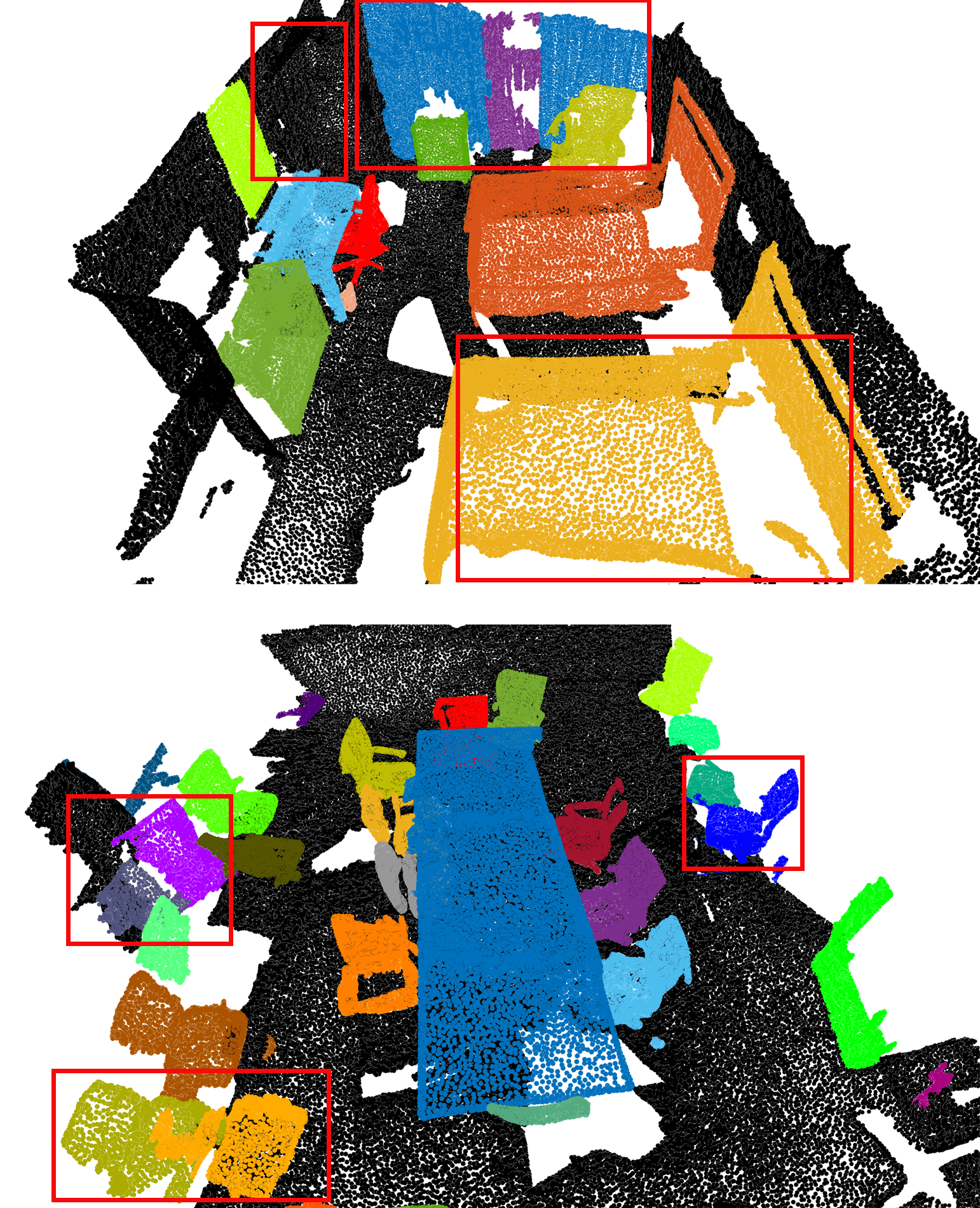}\\
    Ground Truth
    \end{minipage}%
    \caption{Visualiztion of instance segmentation results on the ScanNetv2 validation set. The red box highlight the key regions. \textbf{Zoom in for best view.}}
    \label{fig:vis}
\end{figure*}

\begin{figure*}[t]
    \centering
    \includegraphics[width=\textwidth]{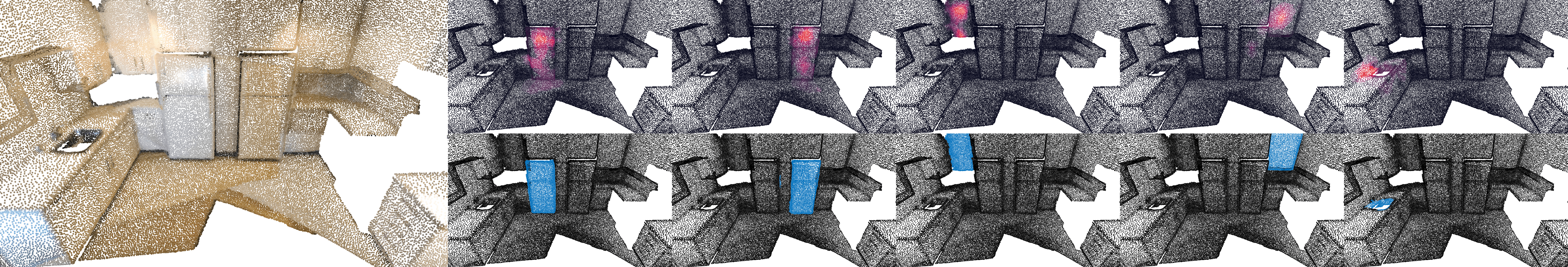}
    \caption{Visualiztion for superpoint cross-attention mechanism. It presents the visualizations of the attention weights in query vectors and corresponding segmentation masks. For the input point cloud of a kitchen scene, the upper row is heat maps, showing the relative attention weights between query vectors and points. The bottom row shows the corresponding mask prediction of each query vector. \textbf{Zoom in for best view.}}
    \label{fig:attn_vis}
\end{figure*}

\begin{table}[!t]
    \centering
    \resizebox{\linewidth}{!}{
    \begin{tabular}{ccc|ccc}
        \toprule
        Superpoint Pooling & Matching Method & Score & mAP & $\text{AP}_{50}$ & $\text{AP}_{25}$ \\
        \midrule
         & mask &  & 34.3 & 54.7 & 72.9 \\
        \checkmark & box & & 49.9 & 68.1 & 78.9 \\
        \checkmark & mask & & 55.0 & 72.4 & 82.5 \\
        \checkmark & mask & \checkmark & \textbf{56.3} & \textbf{73.9} & \textbf{82.9} \\
        \bottomrule
    \end{tabular}}
    \caption{Components analysis on ScanNetv2 validation set.}
    \label{tab:key component}
\end{table}

\begin{table}[!tb]
    \centering
    \resizebox{\linewidth}{!}{
    \begin{tabular}{cccc|ccc}
        \toprule
        \multicolumn{1}{c}{\multirow{2}{*}{\begin{tabular}[c]{@{}c@{}}Iterative\\ prediction\end{tabular}}} & \multicolumn{1}{c}{\multirow{2}{*}{\begin{tabular}[c]{@{}c@{}}Attention\\ mask\end{tabular}}} &
        \multicolumn{1}{c}{\multirow{2}{*}{\begin{tabular}[c]{@{}c@{}}Position\\ encoding\end{tabular}}} &
        \multicolumn{1}{c|}{\multirow{2}{*}{\begin{tabular}[c]{@{}c@{}}Cross-attention\\ first\end{tabular}}} &
        \multirow{2}{*}{mAP} & \multirow{2}{*}{$\text{AP}_{50}$} & \multirow{2}{*}{$\text{AP}_{25}$} \\
        \multicolumn{1}{c}{} & \multicolumn{1}{c}{} & \multicolumn{1}{c}{} & \multicolumn{1}{c|}{}& \multicolumn{1}{c}{} & \multicolumn{1}{c}{} & \multicolumn{1}{c}{} \\
        \midrule
         & & & & 51.0 & 69.6 & 79.8 \\
        \checkmark & & & & 52.5 & 71.4 & 81.6  \\
        \checkmark & \checkmark & & & 56.0 & 73.3 & 82.6 \\
        \checkmark & \checkmark & \checkmark & & 55.6 & 72.7 & 82.0 \\
        \checkmark & \checkmark & & \checkmark & \textbf{56.3} & \textbf{73.9} & \textbf{82.9}\\
        \bottomrule
    \end{tabular}}
    \caption{Ablation study on the architecture of transformer.}
    \label{tab:transformer component}
\end{table}

\paragraph{Number of Queries and Layers.} Table \ref{tab:layer query} presents the selection of the number of query vectors and transformer decoder layers. The results show that too less or too many layers will cause a reduction in performance. Interestingly, we observe some performance improvement when using 400 query vectors compared to 200/100 ones and performance only saturates when the number rises to 800. It may be due to the fact that the number of instances in a 3D scene is usually more than the number of instances in the common 2D dataset.

\begin{table}[!t]
    \centering
    \begin{tabular}{cc|ccc}
        \toprule
        Layer & Query & mAP & $\text{AP}_{50}$ & $\text{AP}_{25}$ \\
        \midrule
        1 & 400 & 49.1 & 66.9 & 79.0 \\
        3 & 400 & 54.7 & 72.3 & 81.5  \\
        6 & 400 & \textbf{56.3} & \textbf{73.9} & 82.9 \\
        12 & 400 & 55.3 & 73.1 & 82.6 \\
        6 & 100 & 54.2 & 72.4 & 82.8 \\
        6 & 200 & 55.2 & 73.3 & 82.4 \\
        6 & 800 & 55.9 & 73.7 & \textbf{83.8} \\
        \bottomrule
    \end{tabular}
    \caption{The performance results of different choices of query vectors and transformer decoder layers.}
    \label{tab:layer query}
\end{table}

\paragraph{The Selection of Mask Loss.} Table \ref{tab:loss} illustrates the performance of the components of mask loss. We observe that only using binary cross-entropy loss or focal \cite{focal} loss will cause much lower performance. Dice loss is indispensable in mask loss. Based on dice loss, adding bce loss or focal loss will improve the total performance. The combination of dice loss and bce loss achieves the best results. 

\subsection{Visualizations}
\paragraph{Qualitative Results.} The visualization of 3D instance segmentation is  shown in Fig. \ref{fig:vis}. Compared to the existing state-of-the-art method, SPFormer correctly segments each instance and produces finer segmentation results.

\paragraph{Cross-Attention Mechanism.} Fig. \ref{fig:attn_vis} visualizes the cross-attention mechanism. For an input point cloud, query vectors attend to the superpoints and highlight the region of interest. Here we propagate the attention weights of superpoints to their own points for visualization. Then query vectors carry the attention information and form the final mask prediction in prediction head.

\begin{table}[!tb]
    \centering
    \begin{tabular}{ccc|ccc}
        \toprule
        Dice       & Focal      & BCE        & mAP           & $\text{AP}_{50}$ & $\text{AP}_{25}$ \\
        \midrule
                   & \checkmark &            & 23.1          & 35.0             & 47.3             \\
                   &            & \checkmark & 35.3          & 52.1             & 68.1             \\
        \checkmark &            &            & 54.8          & 72.8             & 82.4             \\
        \checkmark & \checkmark &            & 55.1          & 73.2             & 82.1             \\
        \checkmark &            & \checkmark & \textbf{56.3} & \textbf{73.9}    & \textbf{82.9}   \\
        \bottomrule
    \end{tabular}
    \caption{Ablation study on the selection of mask loss.}
    \label{tab:loss}
\end{table}

\section{Conclusion}
\label{sec:conclu}
In this paper, we propose a novel end-to-end two-stage framework (SPFormer) for 3D instance segmentation. It can be considered as a combination of proposal-based method and grouping-based method. SPFormer with a novel hybrid pipeline groups bottom-up potential features from point clouds into superpoints and proposes instances by query vectors as a top-down pipeline. SPFormer achieves state-of-the-art on both ScanNetv2 and S3DIS benchmarks, and retains fast inference speed.

\section*{Acknowledgments}
This paper is partially supported by the following grants: National Natural Science Foundation of China (61972163, U1801262), Natural Science Foundation of Guangdong Province (2022A1515011555), National Key R\&D Program of China (2022YFB4500600), Guangdong Provincial Key Laboratory of Human Digital Twin (2022B1212010004) and Pazhou Lab, Guangzhou, 510330, China. 

\bibliography{aaai23}

\end{document}